\pdfoutput=1

\documentclass[runningheads]{llncs}

\usepackage{eccv}

\usepackage{eccvabbrv}

\usepackage{graphicx}
\usepackage{subcaption}       % subfigures (used by ECCV example as well)
\usepackage{booktabs}         % professional tables
\usepackage{amsmath}          % math environments
\usepackage{amssymb}          % for \checkmark
\usepackage{multirow}         % multi-row cells in tables
\usepackage{microtype}        % microtypography
\DeclareRobustCommand{\vec}[1]{\mathbf{#1}}

\usepackage[accsupp]{axessibility}
\usepackage{xcolor}
\usepackage{hyperref}

\usepackage{orcidlink}

\begin{document}

\title{Automated Goldsmith's Mark Retrieval in Silverware}

\titlerunning{Maker's Mark Retrieval}

\author{Atmik Tiwari\inst{1} \and
        Vincent Christlein\inst{1} \and
        Mark Fichtner\inst{2} \and
        Freya Gohlke\inst{2} \and
        Birgit Sch\"{u}bel\inst{2} \and
        Theresa Witting\inst{2} \and
        Heike Zech\inst{2} \and
        Mathias Zinnen\inst{1}}

\authorrunning{A.~Tiwari et al.}

\institute{Pattern Recognition Lab, FAU Erlangen-N\"{u}rnberg \and
           Germanisches Nationalmuseum N\"{u}rnberg}

\maketitle

% TODO: replace the placeholder below with the real proceedings DOI before uploading.
\newcommand{\springerdoi}{10.1007/XXX-X-XXX-XXXXX-X_XX}
\renewcommand{\thefootnote}{}
\footnotetext{This version of the contribution has been accepted for publication,
after peer review, but is not the Version of Record and does not reflect
post-acceptance improvements, or any corrections. The Version of Record is
available online at: \url{https://doi.org/\springerdoi}. Use of this Accepted
Version is subject to the publisher's Accepted Manuscript terms of use
\url{https://www.springernature.com/gp/open-research/policies/accepted-manuscript-terms}.}
\renewcommand{\thefootnote}{\arabic{footnote}}

\begin{abstract}
For art historians, goldsmith marks play a critical role
in the identification and dating of artifacts. 
In practice, experts must manually compare a query mark against hundreds of documented examples, a process that is both tedious and highly dependent on specialist knowledge. 
To address this, we present an AI-assisted retrieval pipeline that combines mark localization with metric-learning fine-tuning across three backbone architectures: 
an ImageNet-pretrained ResNet-50, a supervised ViT-S/16, and a self-supervised DINOv2 ViT-S/14. 
We conduct a systematic evaluation of cropping strategies, where we measure the impact of no cropping, manual ground-truth cropping, and learned detection-based cropping, and assess their interaction with each backbone.
Our strongest configuration, DINOv2 ViT-S/14 with manual crop and metric-learning fine-tuning, achieves an mAP of 62.63\,\% and a Top-1 accuracy of 73.74\,\%. 
Our experiments show that self-supervised pretraining and mark localization are the two most impactful factors, with learned cropping recovering the majority of the gain from manual cropping without requiring ground-truth annotations at inference time.
To enable reproducibility and adoption in the digital humanities, we release our manually annotated dataset and codebase, and deploy the system via a public web interface.

\end{abstract}

\section{Introduction}
\label{sec:intro}

\begin{figure}[tb]
    \centering
    \begin{subfigure}[t]{.49\linewidth}
        \includegraphics[width=\linewidth]{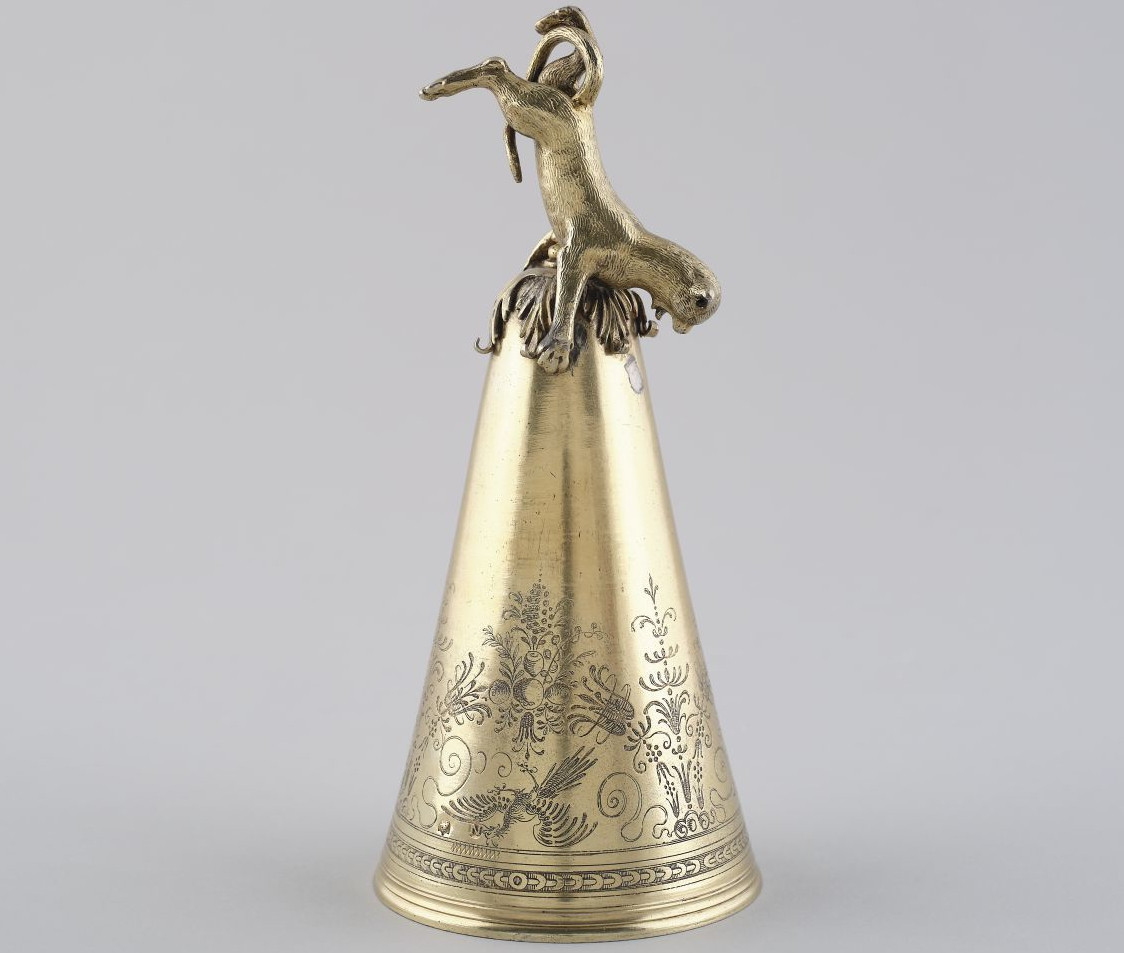}
        \caption{}
        \label{fig:obj}
    \end{subfigure}\hfill%
    \begin{subfigure}[t]{.49\linewidth}
       \includegraphics[width=\linewidth]{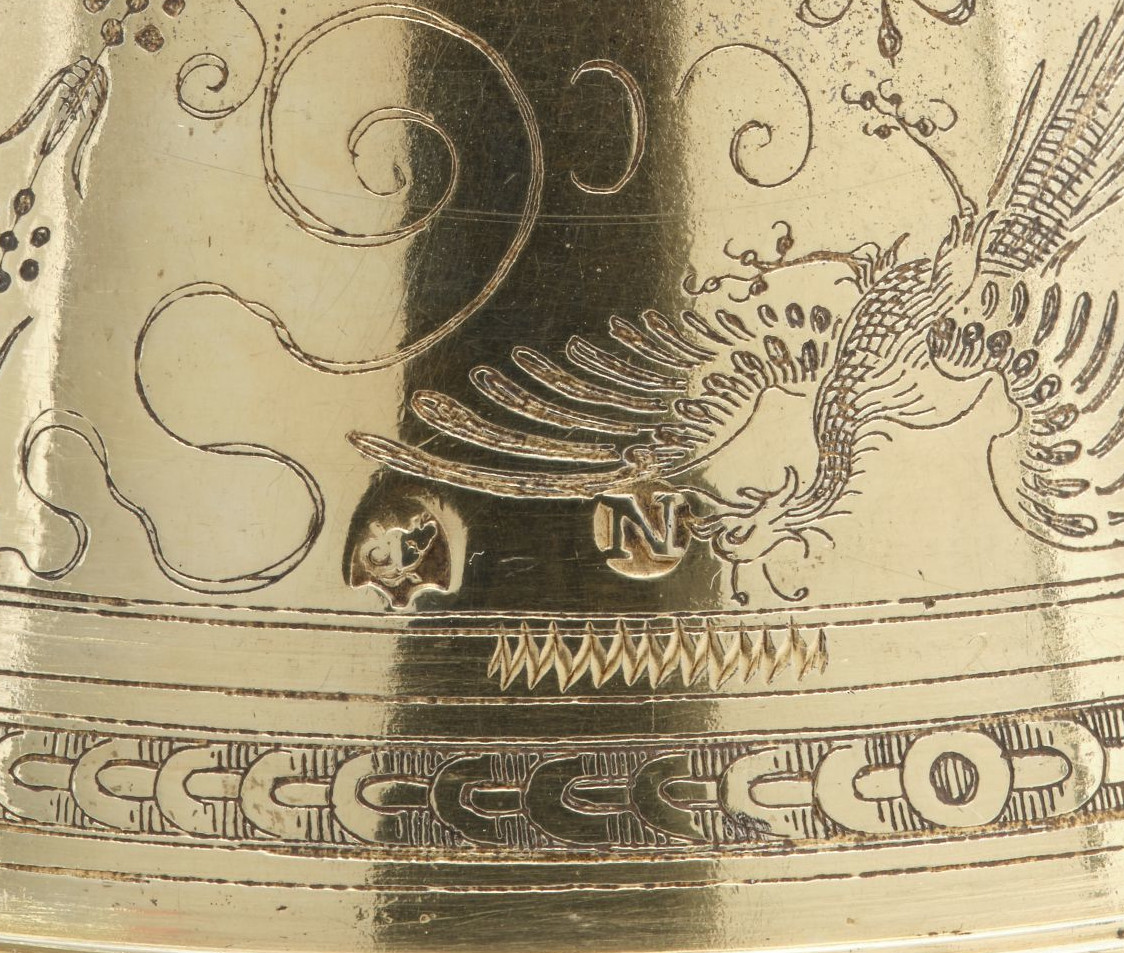}
       \caption{}
       \label{fig:closeup}
    \end{subfigure}
    \caption{Illustration of a hall and a maker's mark on a stirrup cup (Nuremberg, 1609/1629). In the full-object view (\cref{fig:obj}), the marks are difficult to discern; \cref{fig:closeup} provides a close-up. 
    Object credits: Sturzbecher, Hans Petzold, Nürnberg, 1609/1629; Inv. HG10063, GNM, Leihgabe der Jobst Friedrich von Tetzel'sche Familienstiftung.}

    \label{fig:mark_examples}
\end{figure}

In the study of historical gold- and silverwork, maker's marks (\emph{Meisterzeichen}) and hallmarks (\emph{Beschauzeichen}) serve as critical visual identifiers. 
A \emph{Meisterzeichen} denotes the producing workshop via initials or symbols, whereas a \emph{Beschauzeichen} serves as a kind of modern quality mark, with these stamps identifying the town and the workshop. 
As physical impressions made by metal stamps, the morphology, and eventual wear or deformation, of these marks provides vital clues for dating objects and verifying provenance.

While these marks are foundational to art historical research, their systematic visual identification remains a severe bottleneck. 
Marks are extremely small relative to the carrier object (\cref{fig:mark_examples}) and are often subjected to uncontrolled lighting, partial striking, deformation or wear of a stamp, and obscuring reflections or patina. 
Because the visual distinctions between different stamps can be remarkably subtle, manual matching against cataloged examples is tedious and requires highly specialized domain expertise. 
By rapidly retrieving visually similar candidates from large historical corpora, AI-assisted methods can transform this workflow and free experts to focus on contextual interpretation rather than manual visual search.

Motivated by this need, the MarKI project\footnote{\url{https://marki.gnm.de/}} at the Germanisches Nationalmuseum (GNM) aims to build an AI-driven retrieval tool for goldsmith marks. 
This initiative builds upon a comprehensive dataset of over 14{,}000 digital records and mark imprints of Nuremberg goldsmiths (1541--1868) and integrates them into a WissKI~\cite{fichtner2023wisski} semantic environment.

This paper presents the computer-vision pipeline underlying this retrieval system.

At its core, the proposed pipeline relies on Content-Based Image Retrieval (CBIR).
As opposed to classic, textual metadata-enabled retrieval, CBIR systems aim to find semantically similar images within a large database by comparing their extracted visual features. 
While early approaches relied on handcrafted descriptors like SIFT~\cite{lowe2004distinctive} or HOG~\cite{dalal2005histograms}, modern approaches leverage feature extraction via deep neural networks to extract high-level representations~\cite{dubey2021decade}.

In the domain of cultural heritage, these modern CBIR methods are increasingly used to provide image-based search gateways into multiple online collections of paintings and artworks. 
Approaches like imgs.ai~\cite{offert2023imgs} and iART~\cite{springstein2021iart} introduce visual search tools based on multi-modal visual features obtained via CLIP-based~\cite{radford2021learning} pre-training.
For provenance research, Lang et al.~\cite{lang2025digital,zinnen2026aiding} suggest a method to crop depictions of artworks in German Sales~\cite{huemer2014german}, a large collection of historical auction catalogs from the nineteenth and early twentieth centuries, and construct a database of image features to enable image-based retrieval within the corpus. 
Other methods for artwork retrieval rely on image composition~\cite{madhu2023icc++}, posture~\cite{schneider2024poses,schneider2025art}, or hand gestures~\cite{bernasconi2022gab} for retrieving similar artworks from predefined corpora. 
Recently, reranking approaches have been proposed to improve retrieved artwork results with respect to specific criteria~\cite{yemelianenko2023learning}. 
Beyond the computation of feature distances, in specialized applications such as iconography recognition or sensory and smell heritage, it was also suggested to use detected objects~\cite{crowley2014state,zinnen2024smelly}, gestures~\cite{zinnen2025recognizing}, or image-level scene classification~\cite{liu2024novel} to query for image similarity with respect to the respective application focus.
These approaches achieve strong performance primarily by leveraging robustly pre-trained backbones without the need for task-specific fine-tuning. 

However, this is not always the case when working with historical artifacts. 
These objects present particular challenges for retrieval due to their fine-grained nature~\cite{wei2021fine}, where even subtle differences can alter their meaning. 
Furthermore, over centuries of handling and storage in collections, the artifacts accumulate physical wear and degradation. 
Deciphering the relevant nuances and distinguishing between highly similar objects requires expert knowledge and judgment, which significantly complicates automated retrieval approaches~\cite{kutt2026iconographic,zhalehpour2019visual}.
While these challenges are very present in goldsmith marks, they also come up in other domains. 
In computational numismatics, image-based methods have been proposed for the classification of ancient coins, addressing intra-class variation caused by wear and die differences~\cite{anwar2020image}.%

Similarly, in our application, visual features obtained from off-the-shelf pre-trained networks are insufficient to connect similar marks due to the distracting visual features of the carrier objects and varying image modalities, such as lighting conditions. 
An approach to overcome this limitation is metric learning ~\cite{xing2002distance}. 
In metric learning, feature extraction networks are specifically trained to embed images into a space where task-defined semantic similarities correspond to geometric proximity. 
Because our target semantics differ from those of classic ImageNet~\cite{russakovsky2015imagenet} pre-training, we apply metric learning to teach models to focus on the marks' morphology while disregarding the surrounding material. 
Triplet-loss-based~\cite{russakovsky2015imagenet} approaches, in particular, have been shown to reduce the semantic gap~\cite{hattab2026semantic,chen2021integrating}, and the combination of triplet loss with batch hard mining further improves both training efficiency and retrieval performance~\cite{hermans2017defense}.

Building on these principles, this paper presents an evaluation of an automated retrieval pipeline specifically designed for goldsmith marks. In our experiments, we evaluate the impact of metric-learning fine-tuning, mark localization, and different pre-training schemes. 
To ensure reproducibility and encourage adoption in other domains, we release the manually annotated dataset of mark photographs used to train and evaluate our method,\footnote{\url{https://doi.org/10.5281/zenodo.21946908}} alongside our complete code base on GitHub.%
\footnote{\url{https://github.com/Atmickk/VISART2026}}
We introduce a retrieval system that achieves a mean Average Precision (mAP) of 62.63\,\%, demonstrating that our pipeline is feasible under the challenging
conditions of worn surfaces and the fine-grained nature of the retrieval
objects. %
The final pipeline is deployed on the museum's infrastructure and publicly accessible via the project's web interface.\footnote{\url{https://marki.gnm.de/de}}

\section{Method and Materials}
\label{sec:methods}
\subsection{Marks Dataset}

\begin{figure}[t]
  \centering
  \includegraphics[width=\linewidth]{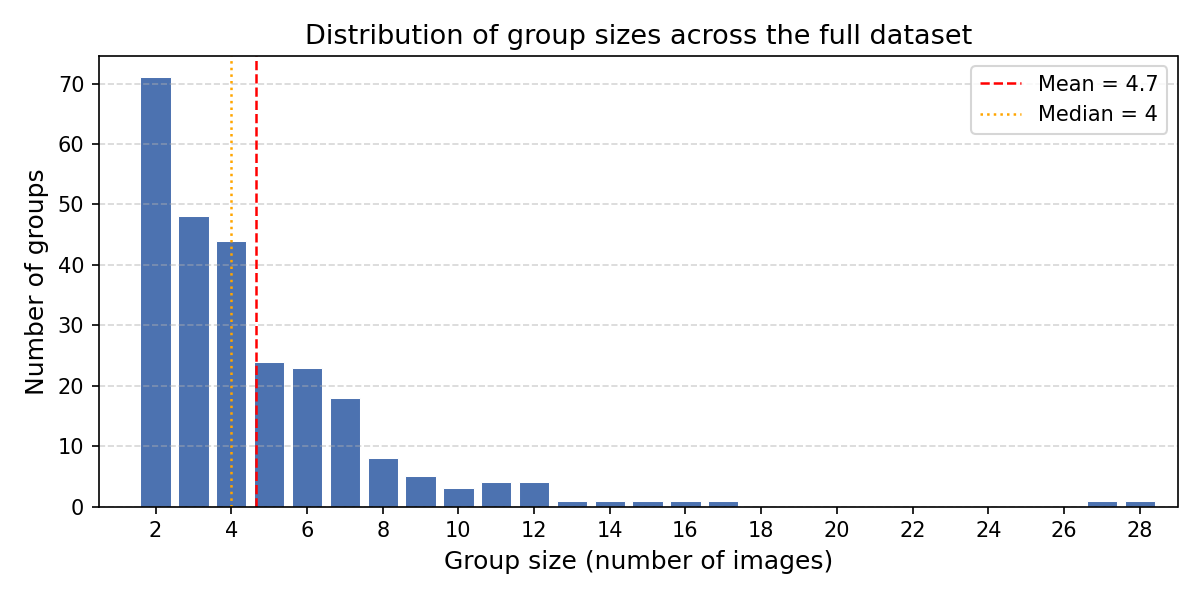}
  \caption{Distribution of group sizes across the full dataset. Each group
    contains marks attributed to the same goldsmith. The distribution is
    strongly right-skewed (mean~$4.6$, dashed; median~$4$, dotted), with most
    groups containing only $2$--$4$ marks and a sparse tail extending to $28$.}
  \label{fig:dataset}
\end{figure}

We retrieved the image files along with the corresponding artist assignments from the public REST endpoint of the project WissKI instance.\footnote{\url{https://ngk.wisski.cloud
}}
To enable mark cropping and to provide ground truth for training a detection model, we manually annotated the positions of all city marks and maker's marks using the CVAT annotation tool~\cite{cvat}.
The resulting dataset comprises 3,608 images, annotated with 1,509 city marks and 1,207 maker's marks.
Due to the provenance of our mark images, the vast majority of the city marks are from Nuremberg, which makes them not uninformative for our retrieval task as currently defined. 
Therefore, we filter them out and restrict our retrieval experiments to the maker's marks. 
In future work, however, we plan to consider the city marks as well, since variations in their appearance can still be used for provenance research. 
In particular, when the dataset is extended to more source collections, the city marks become more meaningful as their origin can no longer be trivially inferred.
The 1,207 maker's marks were attributed to 259 distinct goldsmiths, forming 259 groups, where each group contains marks by the same goldsmith. 
The number of samples per goldsmith varies between 2 and 28, with a mean group size of 4.6. \Cref{fig:dataset} illustrates the distribution of group sizes across the dataset.

For training and evaluating our methods, we split the dataset into 959 training, 102 validation, and 146 test images.
To prevent test data leakage, we performed this split at the group level and ensured that all marks struck with the same physical stamp remain within the same split. 
This means that groups of positive matches used for retrieval evaluation are never shared across the training and testing sets.

\subsection{Mark Retrieval}
\label{subsec:retrieval}

\begin{figure}[tb]
    \centering
    \includegraphics[width=\linewidth]{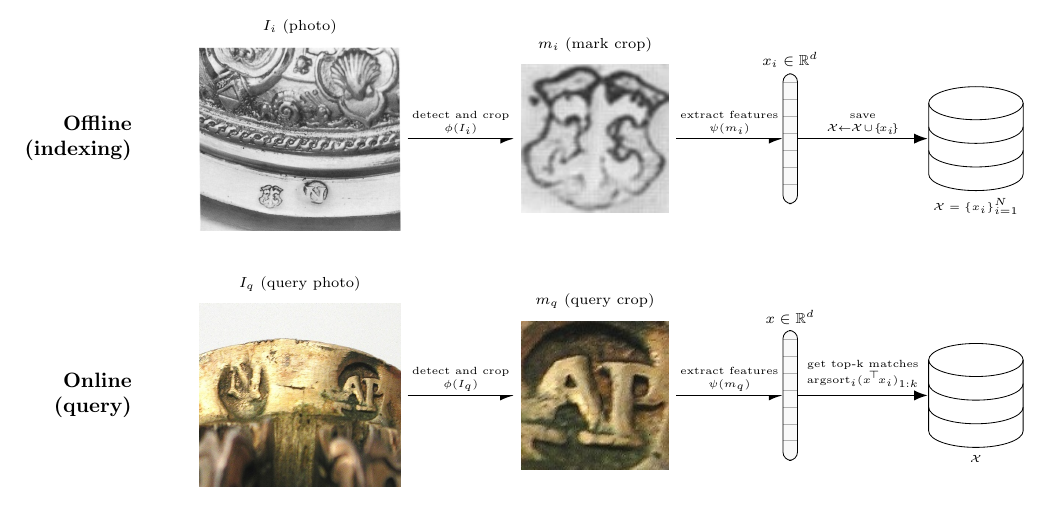}
    \caption{Two-stage retrieval pipeline for maker's marks. Offline, mark crops $m_i=\phi(I_i)$ are embedded as $x_i=\psi(m_i)\in\mathbb{R}^d$ and stored in a feature database $\mathcal{X}$. Online, a query mark is embedded as $x$ and used to retrieve the top-$k$ most similar entries in $\mathcal{X}$, yielding candidate matches likely produced by the same stamp.}
    \label{fig:retrieval}
\end{figure}

Given an object photograph $I$ (\eg, a silverware piece containing one or multiple marks), our goal is to retrieve database entries whose marks are visually most similar to a query mark, ideally corresponding to impressions produced by the same physical stamp. 
We formulate this as nearest-neighbor retrieval in a learned feature space.
Conceptually, this means translating each image into a feature vector where visually similar marks receive similar numeric representations.
An overview of the full pipeline is shown in \cref{fig:retrieval}.

\subsubsection{Offline Indexing.}
In practical terms, this step processes the known museum catalog in advance. 
For each database image $I_i$, we obtain a mark image $m_i$ via a cropping function $\phi$, \ie, $m_i=\phi(I_i)$. The cropped mark is then mapped to a $d$-dimensional feature vector using a feature extractor $\psi$, yielding
\begin{equation}
    x_i = \psi(m_i) \in \mathbb{R}^d.
\end{equation}
All feature vectors are stored in a feature database $\mathcal{X}=\{x_i\}_{i=1}^{N}$, where each $x_i$ retains a reference to its underlying object record and associated metadata (\eg, artist attribution).

\subsubsection{Online Querying.}
When a user searches for a newly photographed mark, the system processes it on the fly.
For a query photograph $I_q$, we analogously compute a query crop $m_q=\phi(I_q)$ and its embedding $x=\psi(m_q)$. We then retrieve the top-$k$ most similar database entries by comparing $x$ to all vectors in $\mathcal{X}$.
In our implementation, we use cosine similarity (dot product with $\ell_2$-normalized embeddings) and return
\begin{equation}
    \mathcal{N}_k(x) = \big(\mathrm{argsort}_i\, x^\top x_i\big)_{1:k},
    \label{eq:topk}
\end{equation}
\ie, the indices of the $k$ highest-scoring feature vectors. 
The corresponding objects (and their mark images) are presented to the user as candidate matches for expert inspection.

\subsection{Mark Detection}
\label{subsec:detection}

This subsection details the cropping function $\phi$ used in \cref{fig:retrieval}. 
Because the marks in our dataset often occupy only a small fraction of the object photographs, we expect cropping the exact mark position to act as a digital magnifying glass that isolates the relevant visual features from surrounding visual noise.
We study three instantiations of $\phi$ that trade off annotation effort, robustness, and deployment practicality.

\subsubsection{(1) No Cropping.}
As a simple baseline, we set $\phi$ to the identity function, \ie, $\phi(I)=I$.
In this setting, feature extraction and retrieval are performed on the full object photograph. 
This requires no mark localization but can be sensitive to background content, scale changes, and the small relative size of marks.

\subsubsection{(2) Manual Cropping.}
As an upper-bound proxy for perfect localization, we use the manually annotated mark bounding boxes available in the dataset and define $\phi(I)=\mathrm{crop}_{\text{gt}}(I)$. 
Retrieval is then performed on the cropped mark regions. 
This isolates the effect of localization quality from the subsequent feature learning and retrieval steps.

\subsubsection{(3) Learned Detection and Cropping.} %
To enable fully automatic retrieval, we learn $\phi$ using an object detector trained to localize marks in photographs.
Concretely, we train a YOLOv11-n~\cite{Jocher_Ultralytics_YOLO26_Unified_2026} detector for 100 epochs to localize bounding boxes for the city marks and maker's marks (including other ambiguous marks).
During inference, $\phi$ applies the detector to an input image $I$, selects the predicted mark box(es), and returns the corresponding crop(s), \ie, $\phi(I)=\mathrm{crop}_{\theta}(I)$.

\subsection{Metric Learning / Feature Extraction}%

To instantiate the feature extractor $\psi$, we employ three pretrained backbone architectures. 
These span three major feature extraction paradigms: a convolutional network (ResNet-50), a supervised Vision Transformer (ViT-S/16), and a self-supervised one (DINOv2 ViT-S/14), which allows attributing performance differences to architecture family and pretraining scheme. Comparing further encoders (\eg, CLIP~\cite{radford2021learning}, DINOv3, or I-JEPA) is left for future work.
In each case, we remove the classification head and attach a projection head, producing 256-dimensional $\ell_2$-normalized embeddings.
We evaluate two configurations for the retrieval task:
(1) Off-the-shelf, where we apply the pretrained model without further
    training and use the resulting feature vectors to compute similarities; and
(2) Fine-tuned, where we fine-tune the network using metric learning to
    obtain feature representations that better capture the specific nuances of
    goldsmith marks.

For the fine-tuned configuration, we use the \texttt{pytorch-metric-learning} library~\cite{Musgrave2020PyTorchML}. Specifically, we employ a triplet margin loss with batch-hard mining and cosine distance. 
The model is trained for up to 100 epochs with early stopping after 25 epochs without improvement. 
All inputs are resized to $224\times224$. 
Data augmentation includes random brightness and contrast adjustment ($p{=}0.5$), hue-saturation-value jitter ($p{=}0.4$), random grayscale ($p{=}0.1$), and Gaussian blur ($p{=}0.3$). 
Gradients are clipped to a maximum norm of 1.0, and $M$-per-class sampling ($M{=}4$) is used with a batch size of 16.

The three backbone-specific configurations are as follows:

\subsubsection{ResNet-50.}
We use an ImageNet-pretrained ResNet-50~\cite{he2016deep} backbone with dropout~\cite{srivastava2014dropout} ($p{=}0.2$) before
the projection head. The model is optimized with Adam~\cite{kingma2014adam} at a learning rate of
$2.737\times10^{-5}$ and weight decay of $2.144\times10^{-5}$, with a triplet
margin of 0.7. A \texttt{ReduceLROnPlateau} scheduler is applied with a reduction
factor of 0.735 and patience of 10. These hyperparameters were obtained via
random search.

\subsubsection{ViT-S/16.}
We use a ViT-Small/16~\cite{dosovitskiyimage} backbone pretrained with supervised ImageNet training
(via \texttt{timm}, \texttt{augreg\_in1k}).

\subsubsection{DINOv2 ViT-S/14.}
We use Meta's self-supervised DINOv2~\cite{oquab2024dinov2} ViT-Small/14 backbone.

Both ViT-based models share the same training configuration. 
They are optimized with AdamW~\cite{loshchilov2018decoupled} using a differential learning rate: $1\times10^{-5}$ for the pretrained backbone and $1\times10^{-3}$ for the randomly initialized projection head. Weight
decay is set to $1\times10^{-4}$ with a triplet margin of 0.2. 
A \texttt{ReduceLROnPlateau} scheduler with reduction factor 0.5 and patience of 10
is applied. No dropout is used.

\section{Results}
\label{sec:results}
We evaluate our proposed pipeline from multiple perspectives. 
First, we establish the overall retrieval performance of our strongest configurations. 
Next, we assess the intermediate mark detection stage, which serves as the foundation for our automated cropping strategy.
To better understand the impact of our metric-learning approach, we then provide a visual analysis of the resulting feature spaces.
Finally, we report the findings of several ablation studies that isolate the individual contributions of the feature extractor architecture, the chosen pretraining strategy, and the cropping methodology on the overall system performance.

\subsection{Overall Retrieval Performance} 
\begin{figure}[tb]
    \centering
    \begin{subfigure}{\linewidth}
        \includegraphics[width=\linewidth]{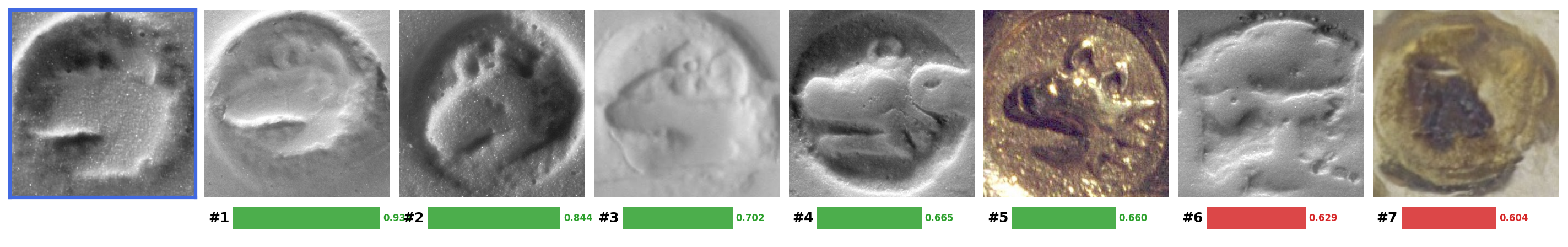}
        \caption{Out of 6 similar images in the database, the model retrieves 5 within the first 7 results.}
        \label{fig:compa}
    \end{subfigure}
    \vspace{4pt}
    \begin{subfigure}{\linewidth}
        \includegraphics[width=\linewidth]{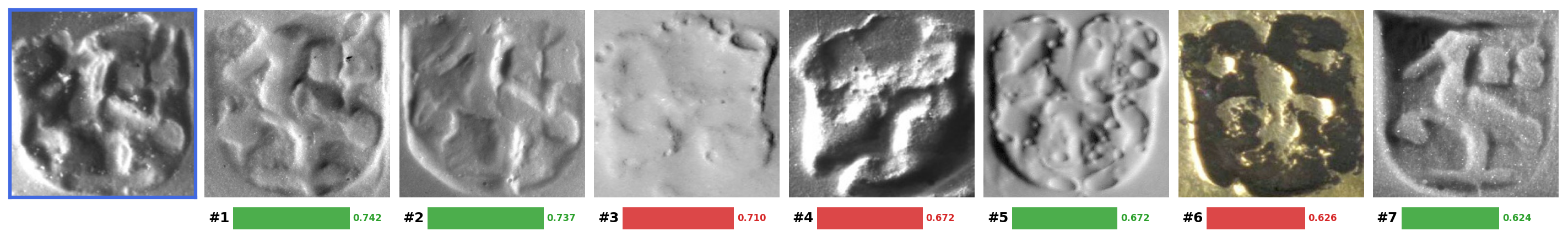}
        \caption{Out of 10 similar images in the database, the model retrieves 4 within the first 7 results.}
        \label{fig:compb}
    \end{subfigure}
    \vspace{4pt}
    \begin{subfigure}{\linewidth}
        \includegraphics[width=\linewidth]{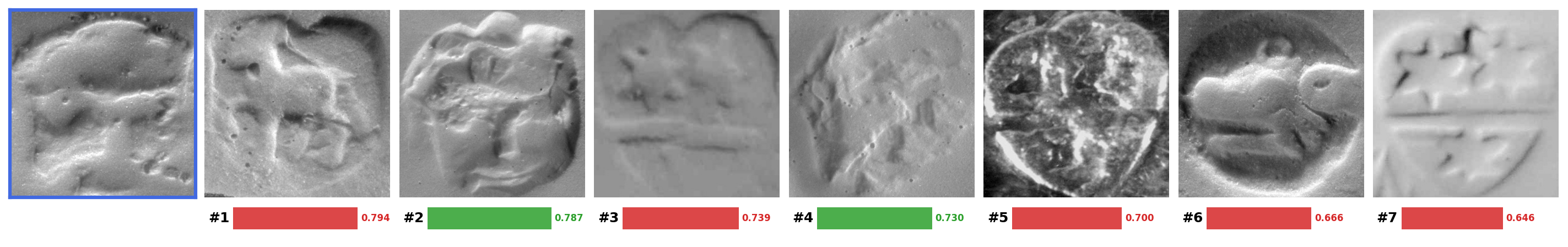}
        \caption{Out of 8 similar images in the database, the model retrieves only 2 within the first 7 results, illustrating a failure case.}
        \label{fig:compc}
    \end{subfigure}
    \caption{Example query results of our strongest retrieval configuration (unseen during training). Each query image (blue border in the left) is followed by the first seven retrieval results, considered most similar by our model. The length of the bars below each result visualizes the similarity between 0 and 1 in feature space. Their color indicates whether the result is a true match (green) or a false positive (red).}
    \label{fig:retrieval_ui}
\end{figure}

\begin{figure}[tbh]
    \centering
    \begin{subfigure}{.48\linewidth}
        \centering
        \includegraphics[width=\linewidth]{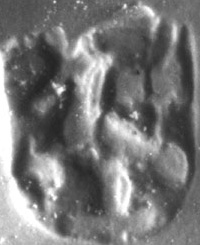}
        \caption{}
        \label{fig:closeupbquery}
    \end{subfigure}%
    \hfill
    \begin{subfigure}{.499\linewidth}
        \centering
        \includegraphics[width=\linewidth]{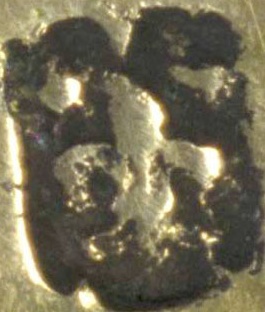}
        \caption{}
        \label{fig:closeupbresult}
    \end{subfigure}
    \caption{Detailed comparison between the second query mark (\cref{fig:closeupbquery}) and its corresponding false positive at rank six (\cref{fig:closeupbresult}). While the overall stamp outline matches the query remarkably well, the internal motifs are similar but visibly distinct.}
    \label{fig:closeupb}
\end{figure}

\begin{figure}[tbh]
        \centering
        \includegraphics[width=.8\linewidth]{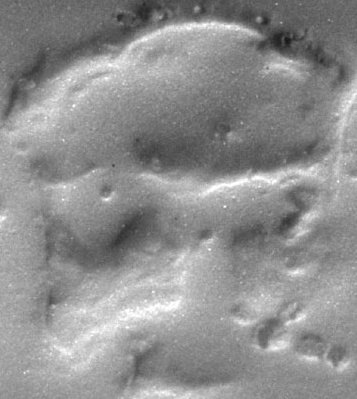}
        \caption{Detailed view of the third query mark. The severe physical wear and deformation illustrate the challenging artifact conditions that lead to retrieval failures.}
    \label{fig:closeupc}
\end{figure}

To evaluate our system, we report mAP and Top-$k$ accuracy (Top-1 and Top-10 in our case). 
These metrics were selected because they effectively evaluate the utility of a retrieval system: mAP tells us how well a system ranks all relevant results, while Top-$k$ accuracy reflects how often the correct result appears within the first $k$ retrieved items, which closely matches real user behavior in retrieval systems.
We treat mAP as our primary metric, as it reflects ranking quality across the full result list, and report Top-1 and Top-10 as intuitive, user-facing complements. Given the modest size of our test set, per-cutoff accuracies should be read as indicative rather than precise.
Larger-scale evaluation would yield more stable estimates.

\Cref{tab:retrieval_combinations} summarizes the retrieval performance across all combinations of cropping strategy $\Phi$ and feature extraction $\Psi$. 
The strongest overall configuration is DINOv2 ViT-S/14 with manual cropping and metric-learning fine-tuning. 
It achieves an mAP of 62.63\,\%, a Top-1 accuracy of 73.74\,\%, and a Top-10 accuracy of 92.69\,\%. 
This means that our retrieval system finds a mark produced by the same goldsmith as the first hit in approximately 74\% of cases, and successfully places a true match within the top 10 results in over 92\% of the cases.

\subsection{Qualitative Analysis}
\Cref{fig:retrieval_ui} shows qualitative examples of retrieved marks for a set of query images from the test set. Neither the queries nor the search space has been seen during training.
In \cref{fig:compa}, the system works optimally.
All five target marks are retrieved within the first five results, despite varying levels of wear, lighting, and deformation. 
Note particularly that the system successfully retrieves the last final mark, which is stamped onto a visually distinct background material compared to the query image.
\Cref{fig:compb} shows a mixed picture: the top two results are correctly retrieved and share clear visual similarities with the query image. 
The subsequent two results originate from another stamp and thus count as false positives. 
A closer inspection of these false positives reveals why the system might conflate them: 
They are heavily eroded, and aside from the general outline that roughly matches that of the query sample, there are hardly any details discernible. 
Even for a human observer without domain expertise, it is difficult to determine whether they originate from the same stamp, which highlights the complexity of the task.
The result on the sixth rank, on the other hand, presents a slightly different false positive. 
\Cref{fig:closeupb} shows a close-up of the query image next to this result. 
While the internal motif resembles that of the query image, a closer look will also allow a layman to distinguish the two. 
However, note that here as well, the overall stamp outline and shape match the query remarkably well.
The third query, illustrated in \cref{fig:compc}, provides an example of a mostly failed retrieval. 
The top-ranked result is immediately a false positive, and only two out of eight true target marks appear at ranks 2 and 4. While the strong wear and deformation on the query sample partly explain this failure (cf. the detailed view in \cref{fig:closeupc}), the example highlights that there is still room for improvement to ensure the model remains robust under such challenging physical conditions. 

\begin{table}[tb]
\caption{Retrieval results for all combinations of cropping $\phi$ and
metric-learning fine-tuning backbone. Trained models are reported as
mean\,$\pm$\,std over 3 random seeds. Best mean per metric in \textbf{bold}.}
\label{tab:retrieval_combinations}
\centering
\scriptsize
\setlength{\tabcolsep}{5pt}
\renewcommand{\arraystretch}{1.6}
\begin{tabular}{lll l ccc}
\toprule
\multicolumn{3}{c}{$\mathbf{\mathrm{\Psi}}$} & \multicolumn{1}{c}{$\mathrm{\Phi}$} \\
\cmidrule(lr){1-3}  \cmidrule(lr){4-4}
\textbf{Backbone} & \textbf{Pretraining} & \textbf{FT} & \textbf{Crop} &
\textbf{mAP (\%)} & \textbf{Top-1 Acc. (\%)} & \textbf{Top-10 Acc. (\%)} \\
\midrule
\multirow{3}{*}{ResNet-50} & \multirow{3}{*}{ImageNet} & \multirow{3}{*}{No} 
  & None  & 17.14 & 15.75 & 62.33 \\
  & & & Learned & 25.83 & 32.88 & 67.12 \\
  & & & Manual & 25.37 & 35.62 & 66.44 \\
\midrule
\multirow{3}{*}{ResNet-50} & \multirow{3}{*}{ImageNet} & \multirow{3}{*}{Yes}
  & None  & $31.56_{\,\pm2.44}$ & $35.16_{\,\pm4.20}$ & $74.53_{\,\pm2.15}$ \\
  & & & Learned & $32.78_{\,\pm3.35}$ & $38.59_{\,\pm6.19}$ & $79.45_{\,\pm1.12}$ \\
  & & & Manual  & $36.87_{\,\pm1.50}$ & $40.64_{\,\pm2.33}$ & $77.40_{\,\pm2.24}$ \\
\midrule
\multirow{3}{*}{ViT-S/16} & \multirow{3}{*}{ImageNet} & \multirow{3}{*}{Yes}
  & None  & $24.75_{\,\pm0.48}$ & $34.70_{\,\pm2.65}$ & $69.18_{\,\pm1.48}$ \\
  & & & Learned & $40.78_{\,\pm4.02}$ & $50.23_{\,\pm5.60}$ & $81.05_{\,\pm4.27}$ \\
  & & & Manual & $47.88_{\,\pm1.57}$ & $56.85_{\,\pm0.97}$ & $87.67_{\,\pm2.01}$ \\
\midrule
\multirow{3}{*}{ViT-S/14} & \multirow{3}{*}{DINOv2} & \multirow{3}{*}{Yes}
  & None  & $35.45_{\,\pm5.22}$ & $46.80_{\,\pm7.04}$ & $81.96_{\,\pm5.19}$ \\
  & & & Learned & $53.29_{\,\pm2.76}$ & $64.84_{\,\pm2.81}$ & $89.50_{\,\pm2.76}$ \\
  & & & Manual & $\mathbf{62.63}_{\,\pm1.72}$ & $\mathbf{73.74}_{\,\pm4.89}$ & $\mathbf{92.69}_{\,\pm1.29}$ \\
\bottomrule
\end{tabular}
\end{table}

\subsection{Mark Detection}
\label{sec:deteval}
\begin{figure}[tbh!]
    \centering
    \begin{subfigure}[b]{0.32\linewidth}
        \includegraphics[width=\linewidth]{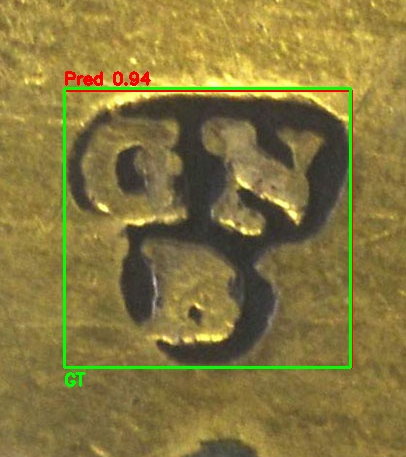}
    \end{subfigure}
    \hfill
    \begin{subfigure}[b]{0.32\linewidth}
        \includegraphics[width=\linewidth]{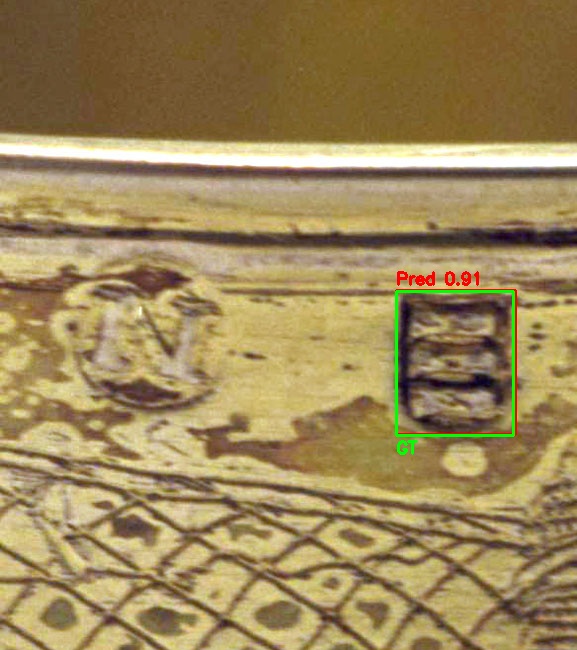}
    \end{subfigure}
    \hfill
    \begin{subfigure}[b]{0.32\linewidth}
        \includegraphics[width=\linewidth]{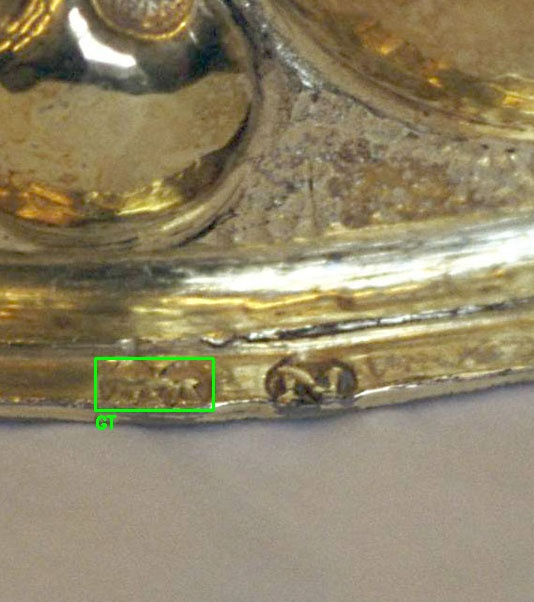}
    \end{subfigure}
    \caption{Qualitative detection results from our trained YOLO model. Green boxes denote ground-truth annotations and red boxes denote model predictions.}
    
    \label{fig:qualitative_results}

\end{figure}
To evaluate our detection stage (learned cropping), we report the mean Average Precision (mAP) at standard Intersection over Union (IoU) thresholds following the COCO evaluation protocol~\cite{lin2014microsoft}. 
Our trained YOLOv11 model achieves an mAP50 of 96.54\,\% and a stricter mAP50-95 of 84.41\,\%. 
We did not observe a notable performance difference between city marks (97.13\,\% AP50) and maker's marks (95.95\,\%). 
\Cref{fig:qualitative_results} shows examples of successful detections and one failure case.

\begin{figure}[htbp!]
    \centering
    \begin{subfigure}[b]{\linewidth}
        \includegraphics[width=\linewidth]{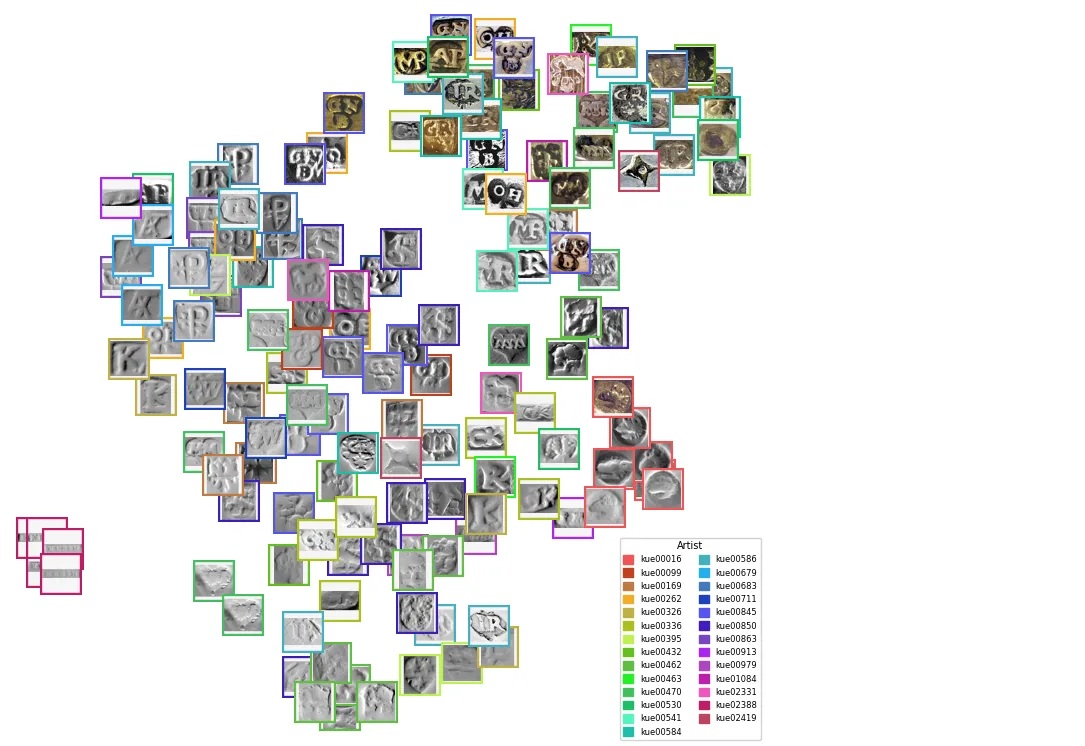}
        \caption{Naïve case}
        \label{fig:umap_naive}
    \end{subfigure}
    \hfill
    \begin{subfigure}[b]{\linewidth}
        \includegraphics[width=\linewidth]{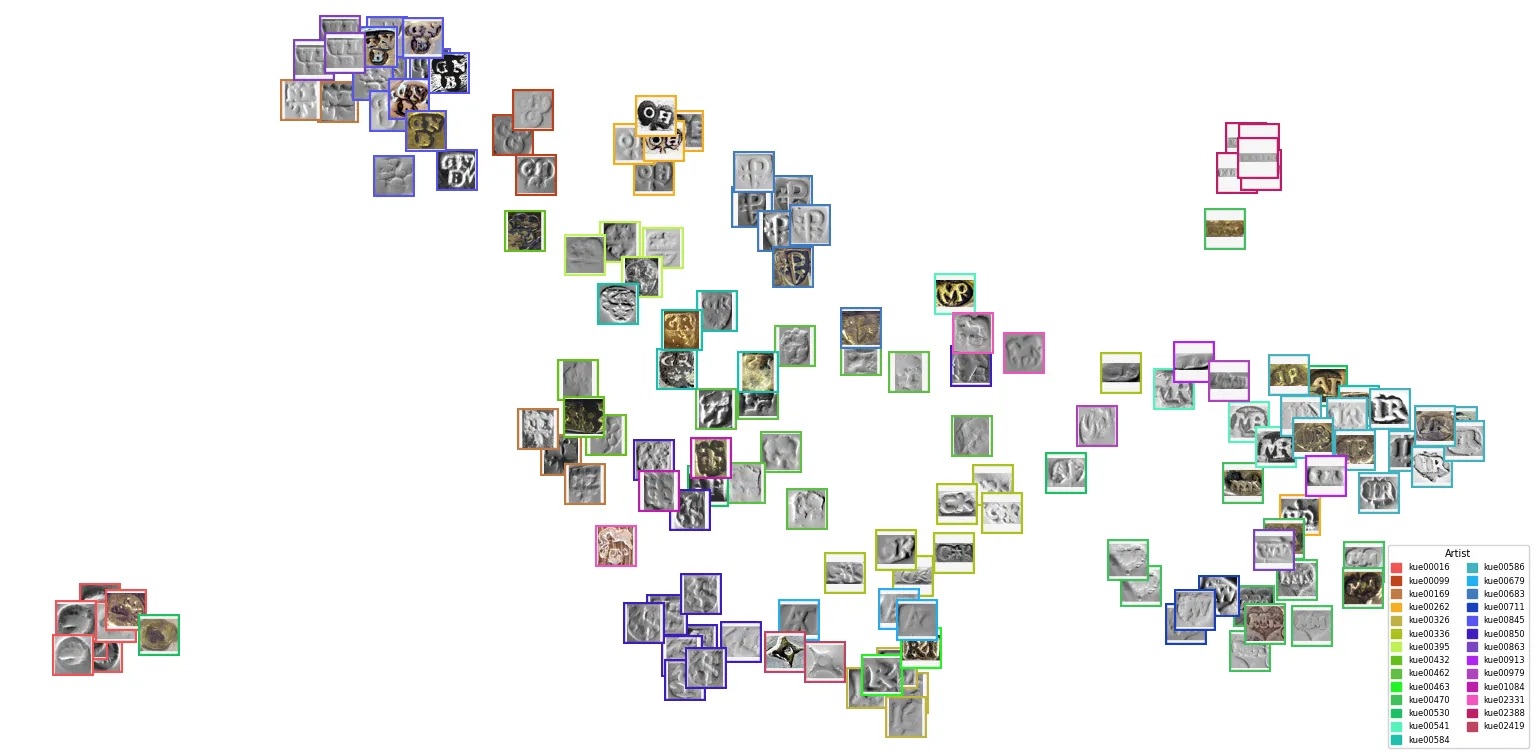}
        \caption{Metric learning}
        \label{fig:umap_metric}
    \end{subfigure}
    \caption{UMAP projection of Off-the-shelf (ResNet50) and metric-learning (DINOv2 ViTS-14) embeddings for the test set (146 manually cropped images)}
    \label{fig:umap_comparison}
\end{figure}
\subsection{Feature Space Analysis}
To further understand the impact of our method, we analyze the feature spaces produced by the extractor $\Psi$ using a UMAP~\cite{mcinnes2018umap} projection.
Based on our quantitative results, we expect the feature embeddings after metric-learning fine-tuning to exhibit better inter-class separability and tighter intra-class clustering. 
The visualization in \cref{fig:umap_comparison} validates this hypothesis. 
In the off-the-shelf baseline (\cref{fig:umap_naive}), it is difficult to visually discern clusters. 
Conversely, after fine-tuning (\cref{fig:umap_metric}), artist-specific clusters are well distinguishable even in the 2D projection.
However, some clusters in our strongest configuration remain overlapping (\cref{fig:umap_metric}). 
This highlights the inherent difficulty of this fine-grained retrieval task and indicates potential for future improvement.

\subsection{Secondary Studies}
\subsubsection{The impact of feature extraction ($\Psi$)}

To isolate the role of the architecture, we compare ResNet-50 and ViT-S/16, both pretrained with ImageNet supervision.
Using the cropped marks (both manual and learned), the attention-based ViT architecture consistently outperforms the convolutional ResNet-50 across all metrics with relative improvements of roughly 30\% in mAP and 40\% in Top-1 accuracy. 
Interestingly, when no cropping is applied, ResNet marginally outperforms the ViT variant. 
This suggests that the attention mechanism is more sensitive to background clutter and benefits more from localization.

\subsubsection{The role of pretraining}

To assess the impact of the selected pretraining strategy, we compare ViT-S/16
(supervised ImageNet) and DINOv2 ViT-S/14 (self-supervised).
These models share a similar ViT-Small architecture and differ only in the patch size.
Across all cropping conditions, the self-supervised DINOv2 backbone consistently outperforms the supervised baseline by a wide margin (15\,\% mAP and 17\,\% Top-1 accuracy). Another key point to note is that these pretraining schemes (self-supervised vs. supervised) also differ in their pretraining data (LVD-142M vs. ImageNet-1k).We therefore cannot fully say that the performance gain is due solely to self-supervision, as it might also be due to the different training data; but nonetheless, DINOv2 clearly performs best and is the backbone we adopt for our retrieval pipeline.

\subsubsection{The impact of cropping ($\Phi$)}
\label{sec:ablation-crop}

Finally, we compare the three cropping strategies: no cropping $\phi(I){=}I$, manual ground-truth cropping $\phi(I){=}\mathrm{crop}_{\mathrm{gt}}(I)$, and learned cropping $\phi(I){=}\mathrm{crop}_{\theta}(I)$. 
As expected, manual cropping consistently achieves the best performance across all fine-tuned backbones.
These gains are particularly pronounced for attention-based models: for ViT-S/16, mAP improves from $24.75\,\%_{\,\pm0.48}$\ (no crop) to $47.88\,\%_{\,\pm1.57}$ (manual crop), and for DINOv2 ViT-S/14 from $35.45\,\%_{\,\pm5.22}$ to $62.63\,\%_{\,\pm1.72}$. 
Interestingly, we observe a contrary behavior in the off-the-shelf baseline without metric-learning fine-tuning. 
Here, the learned crop even achieves a marginally better mAP than the manual crop ($25.83\,\%$ vs.\ $25.37\,\%$). 
We hypothesize that without a task-specific learning signal, the models focus on relatively arbitrary image features, negating the benefit of exact cropping.
The findings furthermore validate the strong detection performance reported in \cref{sec:deteval}. 
Particularly in the ResNet-50 case, learned cropping achieves an mAP of $32.78\,\%_{\,\pm3.35\,\%}$, only marginally below manual crop ($36.87\,\%_{\,\pm1.50}$), while requiring no ground-truth bounding box annotations at inference time.

\section{Conclusion}

In this paper, we presented an automated, two-stage retrieval pipeline designed to identify similar goldsmith marks in challenging real-world museum photography. 
Our experimental evaluations demonstrated that while this fine-grained retrieval task remains difficult, the performance can be substantially improved through a combination of metric-learning fine-tuning and precise localization. Additionally, we found that DINOv2 outperforms standard supervised representations, though we do not fully establish whether this is due to self-supervision alone or also to the difference in training data; in future work, we aim to pinpoint the actual cause. 
Our learned cropping strategy via YOLOv11 successfully recovered the majority of performance gains associated with manual bounding boxes. 
Crop detection can thus enable a fully automated pipeline that requires no ground-truth annotations at inference time.

Despite these advances, some limitations need to be acknowledged.
Our feature space analysis indicates that overlapping clusters remain. 
This shows that physical degradation and the long-tailed distribution of historical samples (with many target groups consisting of only two marks) are still open challenges. 
Furthermore, our current evaluation is based on a relatively small dataset of 146 query images. 
While the method achieves strong performance on this small test set, the current scale frames the proposed pipeline as a pilot study. 
It successfully demonstrates general feasibility but is not yet of real use to provenance researchers who are typically interested in tracing object transitions across multiple collections.
However, the method's efficacy on large-scale, multi-institutional datasets remains to be empirically verified. 
Accordingly, integrating photographs of gold- and silverwork (and the respective marks) from multiple collections presents a crucial next step toward developing a robust, practical tool for art historians and provenance researchers.
Methodologically, it could be explored whether more recent feature encoders such as CLIP~\cite{radford2021learning}, DINOv3~\cite{simeoni2025dinov3}, and I-JEPA~\cite{assran2023self}, or metric learning objectives beyond the triplet loss used in our study, can improve retrieval quality. 

As this work is part of an ongoing, publicly funded research project, we plan to continuously refine the retrieval system and deploy our best-performing models via the project's public web interface. 
We hope that this will eventually enable domain experts to effectively navigate and analyze large corpora of silversmith objects.

More broadly, this retrieval pipeline might also serve as a foundation for more complex analyses of historical gold- and silverwork. 
By accurately retrieving marks struck with the same physical stamp, the system paves the way for a detailed downstream investigation of stamp morphology, \eg by analyzing abrasion, which are crucial for forgery detection and the relative dating of historical artifacts.

\section*{Acknowledgements}
This work was carried out within the MarKI project, funded by the German Federal Ministry of Research, Technology and Space (BMFTR) under the \mbox{DATIpilot} programme (grant no. 03DPS1085). 
MarKI is located at the Germanisches Nationalmuseum (GNM), which we thank for providing access to its collection. 
The underlying dataset builds on the research data of a ten-year, DFG-funded project on Nuremberg goldsmiths' art (1541–1868; DFG project no. 5309288), curated within a WissKI research infrastructure.
We further thank the SODa (Sammlungen, Objekte, Datenkompetenz) project team for their support and advice. 

\clearpage
\bibliographystyle{splncs04}
\bibliography{main}

\end{document}